\def\year{2019}\relax
\documentclass[letterpaper]{article} 
\usepackage{aaai19}  
\usepackage{times}  
\usepackage{helvet}  
\usepackage{courier}  
\usepackage{url}  
\usepackage{graphicx}  
\usepackage{amsfonts}
\usepackage{amsmath}
\usepackage[usenames, dvipsnames, table]{xcolor}
\definecolor{col1}{RGB}{234,209,220}
\definecolor{col2}{RGB}{217, 211, 0}

\title{Depth Prediction Without the Sensors: Leveraging Structure for Unsupervised Learning from Monocular Videos}
\author{Vincent Casser\thanks{Work done while at Google Brain.}\textsuperscript{1} \hspace{20pt} Soeren Pirk \hspace{20pt} Reza Mahjourian\textsuperscript{2} \hspace{20pt} Anelia Angelova\\\\
Google Brain\\
\textsuperscript{1} Institute for Applied Computational Science, Harvard University; Google Brain\\
\textsuperscript{2} University of Texas at Austin; Google Brain\\
vcasser@g.harvard.edu, \{pirk, rezama, anelia\}@google.com
}

\begin{document}
\maketitle

\begin{abstract}
Learning to predict scene depth from RGB inputs is a challenging task both for indoor and outdoor robot navigation. In this work we address unsupervised learning of scene depth and robot ego-motion where  supervision is provided by monocular videos, as cameras are the cheapest, least restrictive and most ubiquitous sensor for robotics.

Previous work in unsupervised image-to-depth learning has established strong baselines in the domain. We propose a novel approach which produces higher quality results, is able to model moving objects and is shown to transfer across data domains, e.g. from outdoors to indoor scenes. The main idea is to introduce geometric structure in the learning process, by modeling the scene and the individual objects; camera ego-motion and object motions are learned from monocular videos as input. Furthermore an online refinement method is introduced to adapt learning on the fly to unknown domains.

The proposed approach outperforms all state-of-the-art approaches, including those that handle motion e.g. through learned flow. Our results are comparable in quality to the ones which used stereo as supervision and significantly improve depth prediction on scenes and datasets which contain a lot of object motion. The approach is of practical relevance, as it allows transfer across environments, by transferring models trained on data collected for robot navigation in urban scenes to indoor navigation settings. The code associated with this paper can be found at \url{https://sites.google.com/view/struct2depth}.
\end{abstract}

\noindent Predicting scene depth from input imagery is important for robot navigation, both for indoors and outdoors settings.  
Supervised dense depth prediction per single image has been very successful with deep neural networks~\cite{eigen2014depth,laina2016deeper,Wang2057designing,Li2017two}, where learned models convincingly outperform those with hand-crafted features~\cite{Ladicky2014Discriminatively,Karsch2017depth}. 
However, supervised learning of scene depth requires expensive depth sensors which may not be readily available in most robotics scenarios and may introduce their own sensor noise. To that end a number of unsupervised image-to-depth methods have been proposed, which demonstrate that unsupervised depth prediction models are more accurate than sensor-supervised ones~\cite{zhou2017unsupervised,garg2016unsupervised}, predominantly due to issues with sensor readings, e.g. missing or noisy sensor values. This research led to a number of improvements in which unsupervised methods have decreased prediction errors significantly, including methods that use stereo~\cite{godard2017monodepth}, or independently trained optical flow models during learning~\cite{wang2018learning}.

\begin{figure}[t]
    \centering
    \includegraphics[width=1.0\linewidth]{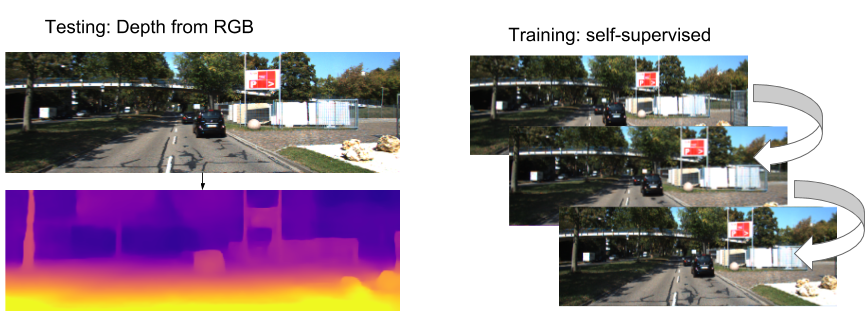}
    \caption{Problem setup: Obtaining scene depth prediction from RGB image input.
   Training is unsupervised and from monocular videos only. No depth sensor supervision is used.}
    \label{fig:first}
\end{figure}

We propose a novel approach that explicitly {\it models 3D motions} of moving objects, together with camera ego-motion, and adapts to new environments by learning with an {\it online refinement} of multiple frames.
With a principled way of handling motion and a newly introduced object size constraint, we are the first to effectively learn from highly dynamic scenes in a monocular setting. Our approach introduces structure in the learning process by representing objects in 3D and modeling motion as SE3 transforms; this is done by fully differentiable operations and is trained from uncalibrated monocular videos. 
Extensive experiments are conducted on two publicly available datasets.
Our algorithm yields significant improvements on both datasets and on both depth and ego-motion estimation, compared to the state-of-the-art; our method is also competitive to models trained with stereo. Furthermore, we evaluate direct domain transfer, by training on one dataset and testing on another, without fine-tuning. We present transfer results across KITTI and Cityscapes, as well as, training on Cityscapes and testing on an indoor Robot Navigation dataset. This demonstrates the method is applicable across domains and that exhaustive data-collection for training may not be needed. The proposed algorithm adapts to new environment and learns to predict depth and ego-motion online. To the best of our knowledge, while online-learning is a well-known concept, we are the first to introduce an online refinement method for domain transfer in this unsupervised learning setting. We do not only show promising results to illustrate this, but also expect the refinement method to be useful in better leveraging temporally and spatially related content during continuous inference. While using online refinement requires additional computation, our main motion model runs at 50 FPS and 30 FPS on a Geforce 1080Ti for batch 4 and 1, respectively, making it real-time capable on several state-of-the-art GPUs.

\section{Previous Work}

Scene depth estimation has been a long standing problem in vision and robotics. Numerous approaches, involving stereo or multi-view depth estimation exist.
Recently a learning-based concept for image-to-depth estimation has emerged fueled by availability of rich feature representations, learned from raw data~\cite{eigen2014depth,laina2016deeper}.
These approaches have shown compelling results as compared to traditional methods~\cite{karsch2014depth}.
Pioneering work in unsupervised image-to-depth learning has been proposed by~\cite{zhou2017unsupervised,garg2016unsupervised} where no depth or ego-motion is needed as supervision. Many subsequent works have improved the initial results in both the monocular setting~\cite{Yang2017unsupervised,yin2018geonet} and when using stereo during training~\cite{godard2017monodepth,ummenhofer2017demon,zhan2018unsupervised,yang2018every}.

However, these methods still fall short in practice because object movements in dynamic scenes are not handled. In these highly dynamic scenes, the abovementioned methods tend to fail as they can not explain object motion. To that end, optical flow models, trained separately, have been used with moderate improvements~\cite{yin2018geonet,yang2018lego,yang2018every}. Our motion model is most aligned to these methods as we similarly use a pre-trained model, but propose to use the geometric structure of the scene and model all objects' motion including camera ego-motion. The refinement method is related to prior work~\cite{bloesch2018codeslam} who use lower dimensional representations to fuse subsequent frames; our work shows that this can be done in the original space to a very good quality.

\begin{figure*}[t]
    \centering        
    \includegraphics[width=0.98\linewidth]{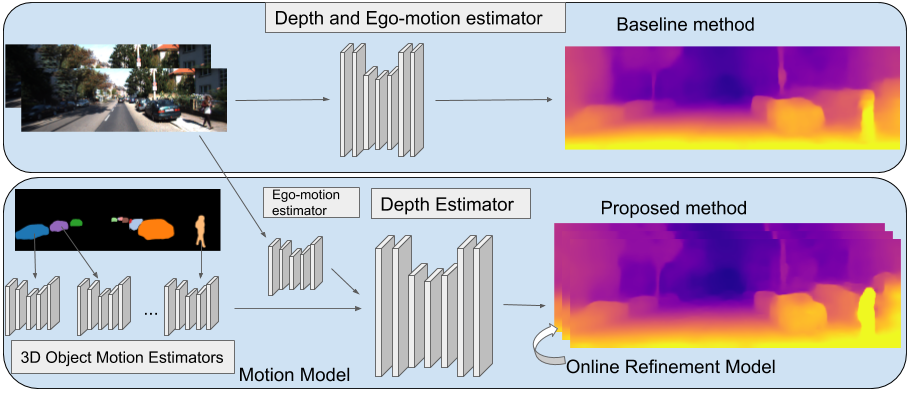}
    \caption{Our method introduces 3D geometry structure during learning by modeling individual objects' motions, ego-motion and scene depth in a principled way. Furthermore, a refinement approach adapts the model on the fly in an online fashion.}
    \label{fig:approach}
\end{figure*}

\section{Main Method}

The main learning setup is unsupervised learning of depth and ego-motion from monocular video~\cite{zhou2017unsupervised}, where the only source of supervision is obtained from the video itself.
We here propose a novel approach which is able to model dynamic scenes by modeling object motion, and that can optionally adapt its learning strategy with an online refinement technique. Note that both ideas are tangential and can be used either separately or jointly. We describe them individually, and demonstrate their individual and joint effectiveness in various experiments.

\subsection{Problem Setup}
The input to the method are sequences of at least three RGB images $(I_1, I_2, I_3)\in \mathbb{R}^{H\times W\times 3}$, as well as camera intrinsics matrix $K\in \mathbb{R}^{3\times 3}$ (we use three for simplicity in all derivations below).
Depth and ego-motion are predicted by learning nonlinear functions, i.e. neural networks. The depth function $\theta:\mathbb{R}^{H\times W\times 3}\rightarrow \mathbb{R}^{H\times W}$ is a fully convolutional encoder-decoder architecture producing a dense depth map $D_i=\theta(I_i)$ from a single RGB frame. The ego-motion network $\psi_E:\mathbb{R}^{2\times H\times W\times 3}\rightarrow \mathbb{R}^{6}$ takes a sequence of two RGB images as input and produces the SE3 transform between the frames, i.e. 6-dimensional transformation vector $E_{1\rightarrow 2}=\psi_E(I_1, I_2)$ of the form $(t_x, t_y, t_z, r_x, r_y, r_z)$, specifying translation and rotation parameters between the frames. Similarly, $E_{2\rightarrow 3}=\psi_E(I_2, I_3)$~\footnote{For convenience the ego-motion network is implemented to obtain two transformations simultaneously from three RGB frames $E_{1\rightarrow 2}, E_{2\rightarrow 3}=\psi_E(I_1, I_2, I_3)$.}.

Using a warping operation of one image to an adjacent one in the sequence, we are able to imagine how a scene would look like from a different camera viewpoint. Since the depth of the scene is available through $\theta(I_i)$, the ego-motion to the next frame $\psi_E$ can translate the scene to the next frame and obtain the next image by projection.
More specifically, with a differentiable image warping operator $\phi(I_i, D_j, E_{i\rightarrow j})\rightarrow \hat{I}_{i\rightarrow j}$, where $\hat{I}_{i\rightarrow j}$ is the reconstructed $j$-th image,  we can warp any source RGB-image $I_i$ into $I_j$ given corresponding depth estimate $D_j$ and an ego-motion estimate $E_{i\rightarrow j}$. In practice, $\phi$ performs the warping by reading from transformed image pixel coordinates, setting $\hat{I}_{i\rightarrow j}^{xy} = I_i^{\hat{x}\hat{y}}$, where $[\hat{x},\hat{y},1]^T=KE_{i\rightarrow j} (D_j^{xy} \cdot K^{-1} [x,y,1]^T)$ are the projected coordinates.
The supervisory signal is then established using a photometric loss comparing the projected scene onto the next frame $\hat{I}_{i\rightarrow j}$ with the actual next frame $I_j$ image in RGB space, for example using a reconstruction loss: $L_{rec}=\min(\Vert \hat{I}_{1\rightarrow 2}-I_2\Vert$.

\subsection{Algorithm Baseline} We establish a strong baseline for our algorithm by following best practices from recent work~\cite{zhou2017unsupervised,godard2018digging}.
The reconstruction loss is computed as the the minimum reconstruction loss between warping from either the previous frame or the next frame into the middle one:
\begin{equation}
L_{rec}=\min(\Vert \hat{I}_{1\rightarrow 2}-I_2\Vert, \Vert \hat{I}_{3\rightarrow 2}-I_2\Vert),
\end{equation}
\noindent proposed by \cite{godard2018digging} to avoid penalization due to significant occlusion/disocclusion effects.
In addition to the reconstruction loss, the baseline uses an SSIM~\cite{wang2004ssim} loss, a depth smoothness loss and applies depth normalization during training, which demonstrated success in prior works ~\cite{zhou2017unsupervised,godard2017monodepth,wang2018learning}. The total loss is applied on $4$ scales ($\alpha_j$ are hyperparameters): \\
\begin{equation}
L=\alpha_1\sum_{i=0}^{3}L^{(i)}_{rec}+\alpha_2L^{(i)}_{ssim}+\alpha_3\frac{1}{2^i}L^{(i)}_{sm}.
\end{equation}
\subsection{Motion Model}
We introduce an object motion model $\psi_M$ which shares the same architecture as the ego-motion network $\psi_E$, but is specialized to predicting motions of individual objects in 3D (Figure~\ref{fig:approach}). Similar to the ego-motion model, it takes an RGB image sequence as input, but this time complemented by pre-computed instance segmentation masks. The motion model is then tasked to learn to predict the transformation vectors per object in 3D space, which creates the observed object appearance in the respective target frame. Thus, computing warped image frames is now not only a single projection based on ego-motion as in prior work~\cite{zhou2017unsupervised}, but a sequence of projections that are then combined appropriately. The static background is generated by a single warp based on $\psi_E$, whereas all segmented objects are then added by their appearance being warped first according to $\psi_E$ and then $\psi_M$. Our approach is conceptually different from prior works which used optical flow for motion in 2D image space~\cite{yin2018geonet} or 3D optical flow~\cite{yang2018every} in that the object motions are explicitly learned in 3D and are available at inference. Our approach not only models objects in 3D but also learns their motion on the fly. This is a principled way of modeling depth independently for the scene and for each individual object.

We define the instance-aligned segmentation masks as $(S_{i,1}, S_{i,2}, S_{i,3}) \in \mathbb{N}^{H\times W}$ per each potential object $i$ in the sequence $(I_1, I_2, I_3)$.
In order to compute ego-motion, object motions are masked out of the images first. More specifically, we define a binary mask for the static scene $O_0(S)=1-\cup_i S_i$, removing all image contents corresponding to potentially moving objects, while $O_j(S)=S_j$ for $j>0$ returns a binary mask only for object $j$. The static scene binary mask is applied to all images in the sequence by element-wise multiplication $\odot$,
before feeding the sequence to the ego-motion model:
$$V=O_0(S_1)\odot O_0(S_2)\odot O_0(S_3)$$
$$E_{1\rightarrow 2}, E_{2\rightarrow 3} = \psi_E(I_1\odot V, I_2\odot V, I_3\odot V)$$

\noindent To model object motion, we first apply the ego-motion estimate to obtain the warped sequences $(\hat{I}_{1\rightarrow 2}, I_2, \hat{I}_{3\rightarrow 2})$ and $(\hat{S}_{1\rightarrow 2}, S_2, \hat{S}_{3\rightarrow 2})$, where the effect of ego-motion has been removed. 
Assuming that depth and ego-motion estimates are correct, misalignments within the image sequence are caused only by moving objects. Outlines of potentially moving objects are provided by an off-the-shelf algorithm \cite{he2017mask} (similar to prior work that use optical flow~\cite{yang2018every} that is not trained on either of the datasets of interest). For every object instance in the image, the object motion estimate $M^{(i)}$ of the $i$-th object is computed as:
\begin{multline}
M^{(i)}_{1\rightarrow 2}, M^{(i)}_{2\rightarrow 3}=\psi_M(\hat{I}_{1\rightarrow 2}\odot O_i(\hat{S}_{1\rightarrow 2}), \\I_2\odot O_i(S_2), \hat{I}_{3\rightarrow 2}\odot O_i(\hat{S}_{3\rightarrow 2}))
\end{multline}
\noindent Note that while $M^{(i)}_{1\rightarrow 2}, M^{(i)}_{2\rightarrow 3}\in \mathbb{R}^6$ represent object motions, they are in fact modeling how the camera would have moved in order to explain the object appearance, rather than the object motion directly. The actual 3D-motion vectors are obtained by tracking the voxel movements before and after the object movement transform in the respective region. Corresponding to these motion estimates, an inverse warping operation is done which moves the objects according to the predicted motions. The final warping result is a combination of the individual warping from moving objects $\hat{I}^{(i)}$, and the ego-motion $\hat{I}$. The full warping $\hat{I}^{(F)}_{1\rightarrow 2}$ is:
\begin{equation}
\hat{I}^{(F)}_{1\rightarrow 2}=\underbrace{\hat{I}_{1\rightarrow 2}\odot V}_{\text{Gradient w.r.t.} \psi_E,\phi}+\sum_{i=1}^{N}\underbrace{\hat{I}^{(i)}_{1\rightarrow 2}\odot O_i(S_2)}_{\text{Gradient w.r.t.} \psi_M, \phi}
\end{equation}

\noindent and the equivalent for $\hat{I}^{(F)}_{3\rightarrow 2}$. In the above, we denote the gradients per each term.
Note that the employed masking ensures that no pixel in the final warping result gets occupied more than once. While there can be regions which are not filled, these are handled implicitly by the minimum loss computation.
Our algorithm will automatically learn individual 3D motion per object which can be used at inference.

\begin{figure*}[t]
    \centering
    {\rotatebox{90}{\hspace{2.6mm} \small Raw}}
    \includegraphics[width=0.24\linewidth]{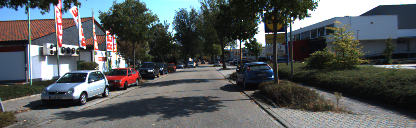}
    \includegraphics[width=0.24\linewidth]{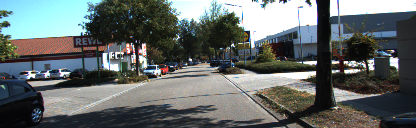}
    \includegraphics[width=0.24\linewidth]{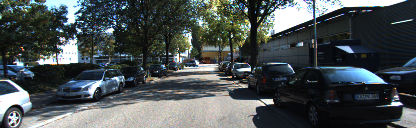}
    \includegraphics[width=0.24\linewidth]{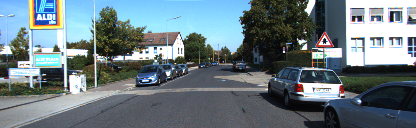}\\
   
    {\rotatebox{90}{\hspace{2.7mm} \small Zhou}}
    \includegraphics[width=0.24\linewidth]{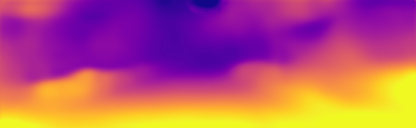}
    \includegraphics[width=0.24\linewidth]{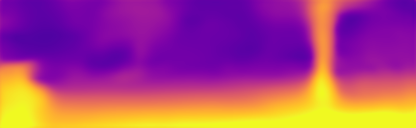}
    \includegraphics[width=0.24\linewidth]{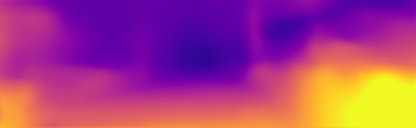}
    \includegraphics[width=0.24\linewidth]{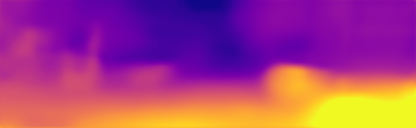}\\
   
    {\rotatebox{90}{\hspace{1.5mm}\small GeoNet}}
    \includegraphics[width=0.24\linewidth]{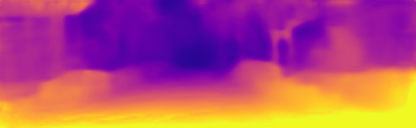}
    \includegraphics[width=0.24\linewidth]{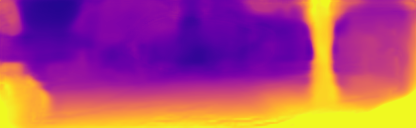}
    \includegraphics[width=0.24\linewidth]{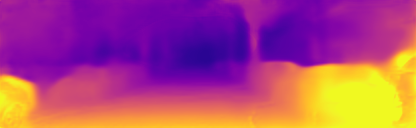}
    \includegraphics[width=0.24\linewidth]{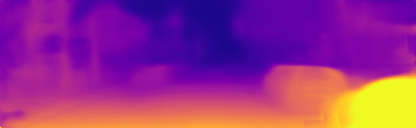}\\
   
    {\rotatebox{90}{\hspace{2.4mm}\small DDVO}}
    \includegraphics[width=0.24\linewidth]{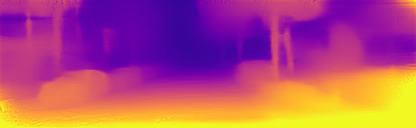}
    \includegraphics[width=0.24\linewidth]{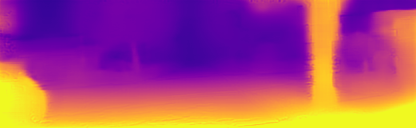}
    \includegraphics[width=0.24\linewidth]{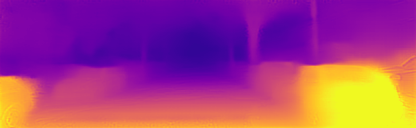}
    \includegraphics[width=0.24\linewidth]{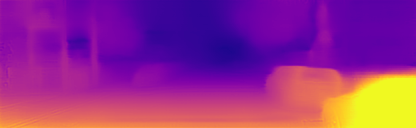}\\
   
    {\rotatebox{90}{\hspace{3mm}\small HMP}}
    \includegraphics[width=0.24\linewidth]{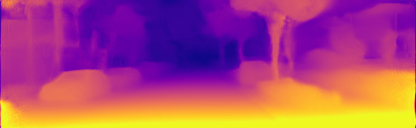}
    \includegraphics[width=0.24\linewidth]{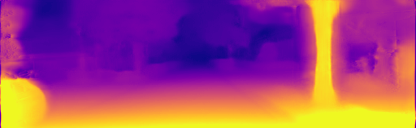}
    \includegraphics[width=0.24\linewidth]{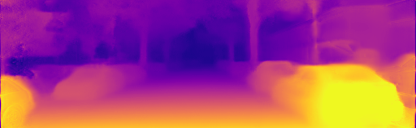}
    \includegraphics[width=0.24\linewidth]{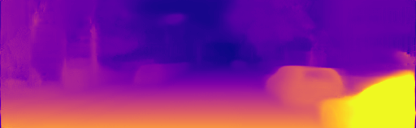}\\
   
    {\rotatebox{90}{\hspace{3mm}\small Ours}}
    \includegraphics[width=0.24\linewidth]{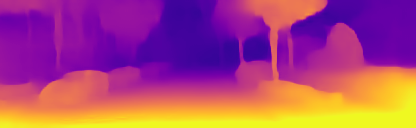}
    \includegraphics[width=0.24\linewidth]{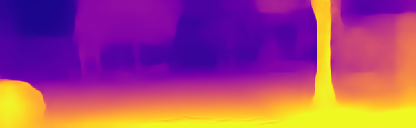}
    \includegraphics[width=0.24\linewidth]{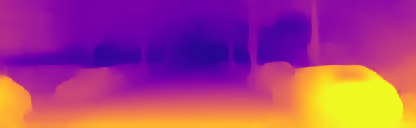}
    \includegraphics[width=0.24\linewidth]{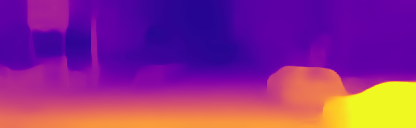}\\
   
    {\rotatebox{90}{\hspace{2.5mm}\small GT}}
    \includegraphics[width=0.24\linewidth]{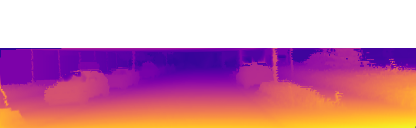}
    \includegraphics[width=0.24\linewidth]{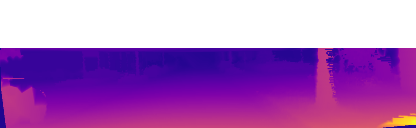}
    \includegraphics[width=0.24\linewidth]{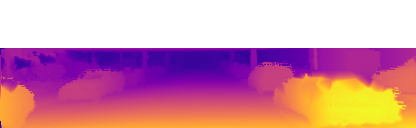}
    \includegraphics[width=0.24\linewidth]{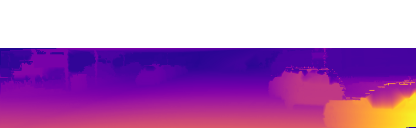}\\

    \caption{Example results of depth estimation compared to the most recent state of the art. Each row shows an input image, depth prediction by competitive methods and ours, and ground truth depth in the last row. KITTI dataset. Best viewed in color.}
    \label{fig:main_kitti}
\end{figure*}

\begin{table*} [h!]
  \centering
  \resizebox{1.0\textwidth}{!}{
  \begin{tabular}{|l|c|c|c||c|c|c|c|c|c|c|}
  \hline
  Method & Supervised? & Motion? & Cap & \cellcolor{col1}Abs Rel & \cellcolor{col1}Sq Rel & \cellcolor{col1}RMSE  & \cellcolor{col1}RMSE log & \cellcolor{col2}$\delta < 1.25 $ & \cellcolor{col2}$\delta < 1.25^{2}$ & \cellcolor{col2}$\delta < 1.25^{3}$\\
  \hline 
  Train set mean & - & - & 80m & 0.361 & 4.826 & 8.102 & 0.377 & 0.638 & 0.804 & 0.894\\
  \hline 
  Eigen \cite{eigen2014depth} Coarse & GT Depth & - & 80m & 0.214 & 1.605 & 6.563 & 0.292 & 0.673 & 0.884 & 0.957\\ 
  Eigen \cite{eigen2014depth} Fine & GT Depth & - & 80m & 0.203 & 1.548 & 6.307 & 0.282 & 0.702 & 0.890 & 0.958\\ 
  Liu \cite{liu2015learning} & GT Depth & - & 80m & 0.201 & 1.584 & 6.471 & 0.273 & 0.68 & 0.898 & 0.967\\
  \hline
  Zhou \cite{zhou2017unsupervised} & - & - & 80m & 0.208 & 1.768 & 6.856 & 0.283 & 0.678 & 0.885 & 0.957 \\
  Yang \cite{Yang2017unsupervised} &- &- & 80m &0.182 &1.481 &6.501 &0.267 &0.725 &0.906 &0.963 \\
  Vid2Depth \cite{mahjourian2018unsupervised} & - & - & 80m & 0.163 & 1.240 & 6.220 & 0.250 & 0.762 & 0.916 & 0.968 \\ 
  LEGO \cite{yang2018lego} &- &M & 80m &0.162 &1.352 &6.276 &0.252 &0.783 &0.921 &0.969 \\
  GeoNet \cite{yin2018geonet} &-  &M & 80m  &0.155 &1.296 &5.857 &0.233 &0.793 &0.931 &0.973 \\
  DDVO \cite{wang2018learning} &- &- & 80m &0.151 &1.257 &5.583 &0.228 &0.810 &0.936 &0.974 \\
  Godard \cite{godard2018digging}  &- &- & 80m  &0.133 &1.158 &5.370 &0.208 &0.841 &0.949 &0.978 \\
  Yang \cite{yang2018every} &- &- & 80m  &0.137 &1.326 &6.232 &0.224 &0.806 &0.927 &0.973 \\
  Yang \cite{yang2018every}  &- &M & 80m  &0.131 &1.254 &6.117 &0.220 &0.826 &0.931 &0.973 \\
  
  \hline
  
  Our (Baseline) & - & - & 80m   &0.1417    &1.1385    &5.5205    &0.2186    &0.8203    &0.9415    &0.9762  \\
  
  Ours (M) & - & M & 80m & 0.1412    & 1.0258    &5.2905   & 0.2153    &0.8160    &0.9452    & 0.9791 \\
    
  Ours (R) & - & - & 80m  &0.1231    &1.4367    &5.3099    &0.2043    & 0.8705    &0.9514    &0.9765 \\

  Ours (M+R) & - & M & 80m  & \textbf{0.1087}    & \textbf{0.8250}    &\textbf{4.7503}    & \textbf{0.1866}    & \textbf{0.8738}    & \textbf{0.9577}    & \textbf{0.9825}  \\
  \hline
  \end{tabular}
  }
  \caption{Evaluation of depth estimation of our method, testing individual contributions of motion and refinement components, and comparing to state-of-the-art monocular methods. The motion column denotes models that explicitly model object motion, while cap specifies the maximum depth cut-off for evaluation purposes in meters. Our results are also close to methods that used stereo (see text). For the purple columns, lower is better, for the yellow ones higher is better. KITTI dataset. }
    \label{tab:kitti_eigen}
\end{table*}

\begin{figure*}[t!]
\centering
\begin{tabular}{@{\hskip 1mm}c@{\hskip 1mm}c@{\hskip 1mm}c@{}}

\end{tabular}
\begin{tabular}{@{\hskip 1mm}c@{\hskip 1mm}c@{\hskip 1mm}c}
\small{Input} & \small{Baseline} & \small{Ours (M)}\\

\includegraphics[width=0.322\linewidth]{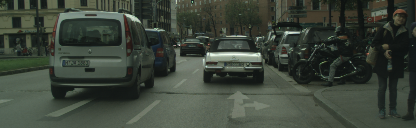} &
\includegraphics[width=0.322\linewidth]{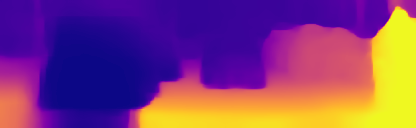} &
\includegraphics[width=0.322\linewidth]{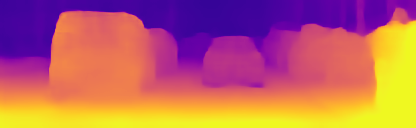} \\

\includegraphics[width=0.322\linewidth]{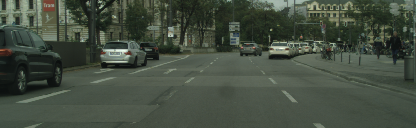} &
\includegraphics[width=0.322\linewidth]{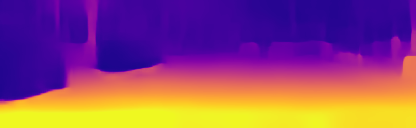} &
\includegraphics[width=0.322\linewidth]{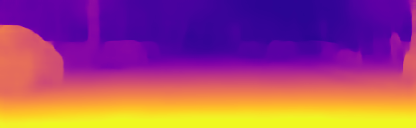} \\

\includegraphics[width=0.322\linewidth]{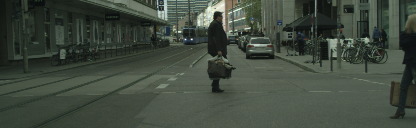} &
\includegraphics[width=0.322\linewidth]{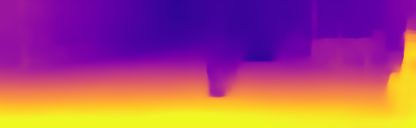} &
\includegraphics[width=0.322\linewidth]{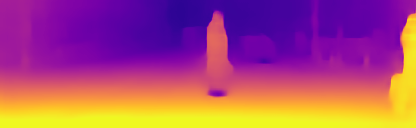} \\
   
\includegraphics[width=0.322\linewidth]{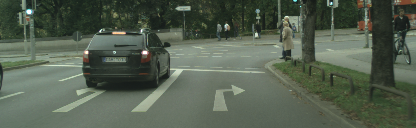} &
\includegraphics[width=0.322\linewidth]{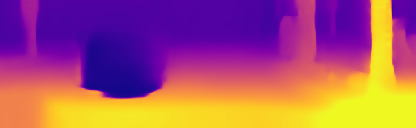} &
\includegraphics[width=0.322\linewidth]{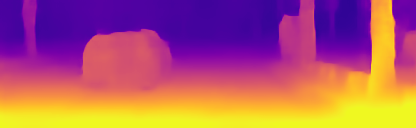} \\

\end{tabular}

    \caption{\textbf{Effect of our motion model (M).} Examples of depth estimation on the challenging Cityscapes dataset, where object motion is highly prevalent. A common failure case for dynamic scenes in monocular methods are objects moving with the camera itself. These objects are projected into infinite depth to lower the photometric error. Our method properly handles this.}
    \label{fig:forward_motion}
\end{figure*}

\begin{figure*} [h!]
    \centering
    \includegraphics[width=1.0\linewidth]{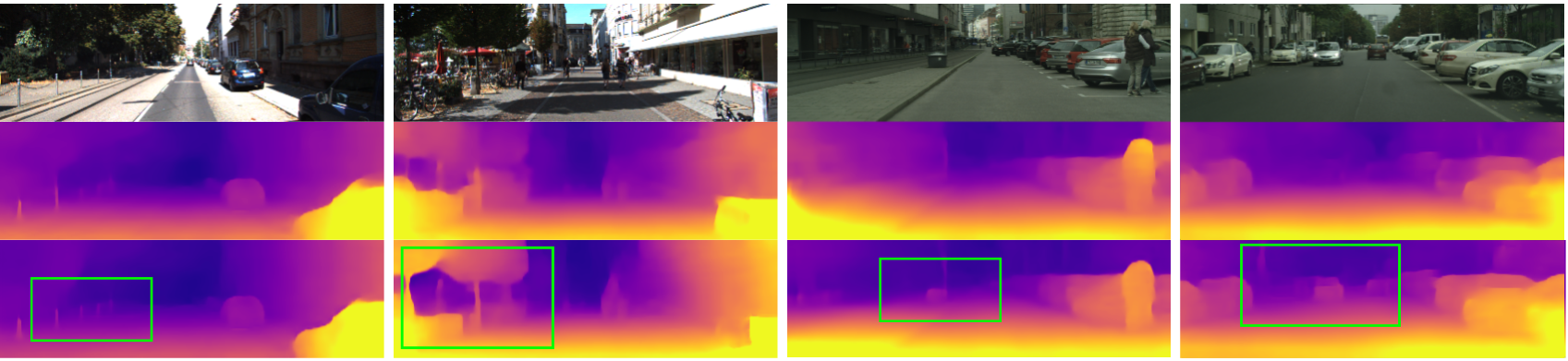}
    \caption{\textbf{Effect of our refinement model (R).} KITTI dataset (left columns), Cityscapes (right columns). Training is done on KITTI for this experiment. Notable improvements are achieved by the refinement model (bottom row), compared to the baseline (middle row), especially for fine structures (leftmost column). The effect is more pronounced on Cityscapes, since the algorithm is applied in zero-shot domain transfer, i.e. without training on Cityscapes itself. }
    \label{fig:kitti_refinement}
\end{figure*}

\subsection{Imposing Object Size Constraints}
A common issue pointed out in previous work is that cars moving in front at roughly the same speed often get projected into infinite depth e.g. \cite{godard2018digging,yang2018every}. This is because the object in front shows no apparent motion, and if the network estimates it as being infinitely far away, the reprojection error is almost reduced to zero which is preferred to the correct case.
Previous work has pointed out this significant limitation \cite{godard2018digging} \cite{yang2018every} \cite{wang2018learning} but offered no solution except for augmenting the training dataset with stereo images. However, stereo is not nearly as widely available as monocular video, which will limit the method's applicability.
Instead, we propose a different way of addressing this problem. The main observation we make is that if the model has no knowledge about object scales, it could explain the same object motion by placing an object very far away and predicting very significant motion, assuming it to be very large, or placing it very close and predicting little motion, assuming it to be very small. Our key idea is to let the model learn objects' scales as part of the training process, thus being able to model objects in 3D.
Assuming a weak prior on the height of certain objects, e.g. a car, we can get an approximate depth estimation for it given its segmentation mask and the camera intrinsics using $D_{\text{approx}}(p;h)\approx f_y \frac{p}{h}$ where $f_y\in \mathbb{R}$ is the focal length, $p\in \mathbb{R}$ our height prior in world units, and $h\in \mathbb{N}$ the height of the respective segmentation blob in pixels. In practice, it is not desirable to estimate such constraints by hand, and the depth prediction scale produced by the network is unknown. Therefore, we let the network learn all constraints simultaneously without requiring additional inputs.
Given the above, we define a loss term on the scale of each object $i$ ($i=1\dots N$). Let $t(i): \mathbb{N}\rightarrow \mathbb{N}$ define a category ID for any object $i$, and $p_j$ be a learnable height prior for each category ID $j$. Let $D$ be a depth map estimation and $S$ the corresponding object outline mask. Then the loss
$$L_{sc}=\sum_{i=1}^{N} \Vert \frac{D \odot O_i(S)}{\overline{D}}-\frac{D_{\text{approx}}(p_{t(i)}; h(O_i(S)))}{\overline{D}}\Vert$$
effectively prevents all segmented objects to degenerate into infinite depth, and forces the network to produce not only a reasonable depth but also matching object motion estimates.
We scale by $\overline{D}$, which is the mean estimated depth of the middle frame, to reduce a potential issue of trivial loss reduction by jointly shrinking priors and the depth prediction range.
To our knowledge this is the first method to address common degenerative cases in a fully monocular training setup in 3D.
Since this constraint is an integral part of the modeling formulation, the motion models are trained with $L_{sc}$ from the beginning. However, we observed that this additional loss can successfully correct wrong depth estimates when applying it to already trained models, in which case it works by correcting depth for moving objects.

\subsection{Test Time Refinement Model}
One advantage of having a single-frame depth estimator is its wide applicability. However, this comes at a cost when running continuous depth estimation on image sequences as consecutive predictions are often misaligned or discontinuous. These are caused by two major issues 1) scaling inconsistencies between neighboring frames, since both our and related models have no sense of global scale, and 2) low temporal consistency of depth predictions.
In this work we contend that fixing the model weights during inference is not required or needed and being able to {\it adapt} the model in an online fashion is advantageous, especially for practical autonomous systems.
More specifically, we propose to keep the model training while performing inference, addressing these concerns by effectively performing online optimization. In doing that, we also show that even with very limited temporal resolution (i.e., three-frame sequences), we can significantly increase the quality of depth predictions both qualitatively and quantitatively. Having this low temporal resolution allows our method to still run on-line in real-time, with a typically negligible delay of a single frame.
The online refinement is run for $N$ steps ($N=20$ for all experiments) which are effectively fine-tuning the model on-the-fly; $N$ determines a good compromise between exploiting the online tuning sufficiently and preventing over-training which can cause artifacts. The online refinement approach can be seamlessly applied to any model including the motion model described above.

\section{Experimental Results}
Extensive experiments have been conducted on  depth estimation, ego-motion estimation and on transfer learning to new environments. We use common metrics and protocols for evaluation adopted by prior methods. With the same standards as in related work, if depth measurements in the groundtruth are invalid or unavailable, they are masked out in the metric computation. We use the following datasets:

\textbf{KITTI dataset (K).} The KITTI dataset~\cite{geiger2013vision} is the main benchmark for evaluating depth and ego-motion prediction. It has LIDAR sensor readings, used for evaluation only. We use standard splits into training, validation and testing, commonly referred to as the `Eigen' split~\cite{eigen2014depth}, and evaluate depth predictions up to a fixed range (80 meters).

\begin{figure}
    \centering
    \includegraphics[width=1.0\linewidth]{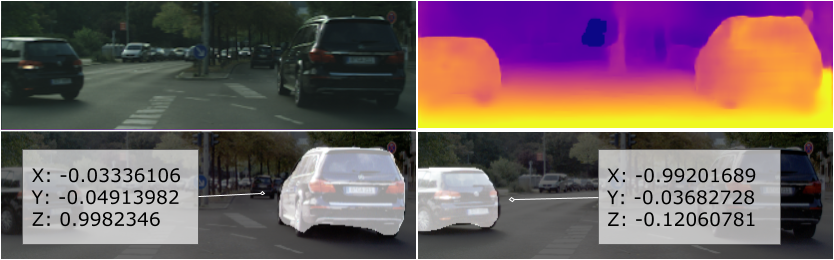}
    \caption{One benefit of our approach is that individual object motion estimates in 3D are produced at inference and the direction and speed of every object in the scene can be obtained. Predicted motion vectors normalized to unit vectors are shown (yaw, pitch, raw are not shown for clarity).}
\label{fig:kitti_vectors}
\end{figure}

\textbf{Cityscapes dataset (C).} The Cityscapes dataset~\cite{cordts2016cityscapes} is another popular and also challenging dataset for autonomous driving. It contains 3250 training and 1250 testing examples which are used in our setup. Of note is that this dataset contains many dynamic scenes with multiple moving objects. We use it for training and for evaluating transfer learning, without fine-tuning.

\begin{table*}
  \centering
  \resizebox{1.0\textwidth}{!}{
  \begin{tabular}{|l|c|c|c|c|c|c|c|}
  \hline
  Method & \cellcolor{col1}Abs Rel & \cellcolor{col1}Sq Rel & \cellcolor{col1}RMSE  & \cellcolor{col1}RMSE log & \cellcolor{col2}$\delta < 1.25 $ & \cellcolor{col2}$\delta < 1.25^{2}$ & \cellcolor{col2}$\delta < 1.25^{3}$\\
  \hline 
  Godard \cite{godard2018digging}* &0.233 &3.533 &7.412 &0.292 &0.700 &0.892 &0.953\\
  \hline
  Our baseline & 0.2054  &  1.6812  &  6.5548  &  0.2751  &  0.6965  &  0.9000  &  0.9612 \\
  Ours (R) & 0.1696  &  1.7083  &  6.0151  &  0.2412  &  0.7840  &  0.9279  &  0.9703\\
  Ours (M) & 0.1876  &  1.3541  &  6.3166  &  0.2641  &  0.7135  &  0.9046  &  0.9667 \\
  Ours (M+R) & \textbf{0.1529}  &  \textbf{1.1087}  &  \textbf{5.5573}  &  \textbf{0.2272}  &  \textbf{0.7956}  &  \textbf{0.9338}  &  \textbf{0.9752}  \\

  \hline
  \end{tabular}
  }
  \caption{Depth prediction results when training on Cityscapes and evaluating on KITTI. Methods marked with an asterik (*) might use a different cropping as the exact parameters were not available.}
    \label{tab:main_city}
\end{table*}

\textbf{Fetch Indoor Navigation dataset.} This dataset is produced by our Fetch robot~\cite{wise2016fetch} collected for the purposes of indoor navigation. We test an even more challenging transfer learning scenario when training on an outdoor navigation dataset, Cityscapes, and testing on the indoor one without fine-tuning. The dataset contains $1,626$ images from a single video sequence, recorded at 8fps.

\subsection{Results on the KITTI Dataset}
Figure~\ref{fig:main_kitti} visualizes the results of our method compared to state-of-the-art methods and Table~\ref{tab:kitti_eigen} shows quantitative results. Both show a notable improvement over the baseline and over previous methods in the literature. With an absolute relative error of $0.1087$, our method is outperforming competitive models that use motion, $0.131$~\cite{yang2018every} and $0.155$~\cite{yin2018geonet}. 
Furthermore, our results, although monocular, are approaching methods which use stereo or a combination of stereo and monocular, e.g.~\cite{godard2017monodepth,Kuznietsov2017semisupervised,yang2018every,godard2018digging}.

\subsubsection{Motion model.}
The main contributions of the motion model are that it is able to learn proper depth for moving objects and  it learns better ego-motion.
Figure~\ref{fig:forward_motion} shows several examples of dynamic scenes from the Cityscapes dataset, which contain many moving objects. We note that our baseline, which is by itself a top performer on KITTI, is failing on moving objects. Our method makes a notable difference both qualitatively (Figure~\ref{fig:forward_motion}) and quantitatively (see Table~\ref{tab:main_city}).   
Another benefit provided by our motion model is that it learns to predict individual object motions. Figure~\ref{fig:kitti_vectors} visualizes the learned motion for individual objects.
See the project webpage for a video which demonstrates depth prediction as well as relative speed estimation which is well aligned with the apparent ego-motion of the video.

\begin{table} [h!]
\centering
\resizebox{0.45\textwidth}{!}{
\begin{tabular}{|l|c|c|}
\hline
Method & Seq. $09$  & Seq. $10$ \\
\hline
Mean Odometry  & 0.032 $\pm 0.026$  & $0.028 \pm 0.023$ \\
ORB-SLAM (short)  & $0.064 \pm 0.141$ & $0.064 \pm 0.130$ \\
Vid2Depth (Mahjourian 2018) & $0.013 \pm 0.010$ & $0.012 \pm 0.011$ \\
Godard (Godard 2018)$\dagger$  & $0.023 \pm 0.013$ & $0.018 \pm 0.014$ \\
Zhou (Zhou 2017)$\dagger$   & $0.021 \pm 0.017$  & $0.020 \pm 0.015$ \\
GeoNet (Yin 2018)  & $ 0.012 \pm 0.007$ & $0.012 \pm 0.009$ \\
ORB-SLAM (full)*  & $0.014 \pm  0.008$ & $0.012 \pm 0.011$ \\
Ours   & $\mathbf{0.011 \pm 0.006}$    & $\mathbf{0.011 \pm 0.010}$\\
\hline
\end{tabular}
}
\caption{Quantitative evaluation of odometry on the KITTI Odometry test sequences. Methods using more information than a set of rolling 3-frames are marked (*). Models that are trained on a different part of the dataset are marked ($\dagger$).}
\label{fig:kitti_vo}
\end{table}
\vspace{-0.2cm}

\begin{figure*}
    \centering   \includegraphics[width=1.0\linewidth]{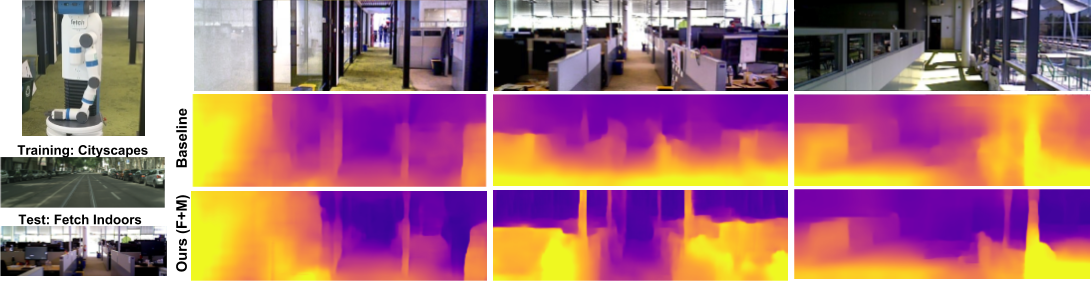}
    \caption{Testing on the Fetch robot Indoor Navigation dataset. The model is trained on the Cityscapes dataset which is outdoors and only tested on the indoors navigation data. As seen our method (bottom row) is able to adapt online and produces much better and visually compelling results than the baseline (middle row) in this challenging transfer setting.}
    \label{fig:fetch}
\end{figure*}

\subsubsection{Refinement model.}
We observe improvements obtained by the refinement model on both KITTI and Cityscapes datasets.
Figure~\ref{fig:kitti_refinement} shows results of the refinement method only as compared to the baseline.
As seen for both evaluating on KITTI or Cityscapes dataset the refinement is helpful in recovering the geometry structure better.
In our results we observe that the refinement model is most helpful when testing across datasets, i.e. in data transfer.

\subsection{Experimental Results on the Cityscapes Dataset}

In this section we evaluate our method on the Cityscapes dataset, where a lot of object motion is present in the training set. Table~\ref{tab:main_city} shows our experimental results when training on the Cityscapes data, and then evaluating on KITTI (without further fine-tuning on KITTI training data).
This experiment clearly demonstrates the benefit of our method as we see significant improvements from 0.205 to 0.153 absolute relative error for the proposed approach, which is particularly impressive in the context of state-of-the-art error of 0.233. It is also seen that improvements are accomplished by both the motion and the refinement model individually and jointly. We note that the significant improvement of the combined model stems from both the appropriate depth learning of many moving objects (Figure~\ref{fig:forward_motion}) enabled by the motion component, and the refinement component that actively refines geometry in the scene (Figure~\ref{fig:kitti_refinement}).

\subsection{Visual Odometry Results}
Table~\ref{fig:kitti_vo} summarizes our ego-motion results, which are conducted by a standard protocol adopted by prior work~\cite{zhou2017unsupervised,godard2018digging} on parts of the KITTI odometry dataset. The total driving sequence lengths tested are 1,702 meters and 918 meters, respectively.
As seen our algorithm performance is the best among the state-of-the-art methods, even compared to ones that use more temporal information, or established methods such as ORB-SLAM. Proper handling of motion is the biggest contributor to improving our ego-motion estimation.

\subsection{Experiments on Fetch Indoor Navigation Dataset}

Finally, we verify the approach in an indoor environment setting, by testing on data collected by the Fetch robot~\cite{wise2016fetch}. This is a particularly challenging transfer learning scenario as training is done on Cityscapes (outdoors) and testing is done on a dataset collected indoors by a different robot platform, representing a significant domain shift between these datasets.
Figure~\ref{fig:fetch} visualizes the results on the Fetch data. Our algorithm produces better and more realistic depth estimates and is able to notably improve the baseline method and successfully adapt to new environments. 
Notably, the algorithm is able to capture well large transparent glass doors and windows and reflective surfaces. We observe that transfer works best if the amount of motion in between frames is somewhat similar. Also, to have additional information available and not lead to degenerate evolution, camera motion should be present. Thus, in a static state, online refinement should not be applied.

\textbf{Implementation details.}
The code is implemented in TensorFlow and publicly available. The input images are resized to $416\times 128$ (with center cropping for Cityscapes).
The experiments are run with: learning rate $0.0002$, L1 reconstruction weight $0.85$, SSIM weight $0.15$, smoothing weight $0.04$, object-motion constraint weight $0.0005$ (although $0.0002$ seems to work better for KITTI), batch size of $4$, L2 weight regularization of $0.05$. We perform on-the-fly augmentation by horizontal flipping during testing.

\section{Conclusions and Future Work}
The method presented in this paper addresses the monocular depth and ego-motion problem by modeling individual objects' motion in 3D. We also propose an online refinement technique which adapts learning on the fly and can transfer to new datasets or environments. The algorithm achieves new state-of-the-art performance on well established benchmarks, and produces higher quality results for dynamic scenes.
In the future, we plan to apply the refinement method over longer sequences so as to incorporate more temporal information. Future work will also focus on full 3D scene reconstruction which is enabled by the proposed depth and ego-motion estimation methods. 

\textbf{Acknowledgements.} We would like to thank Ayzaan Wahid for helping us with data collection.

\fontsize{9.5pt}{10.5pt}
\selectfont

\bibliographystyle{aaai}
\bibliography{main}
\bibliographystyle{aaai}

\end{document}